\documentclass{article}

\usepackage[final,nonatbib]{neurips_2019}

\usepackage[utf8]{inputenc} 
\usepackage[T1]{fontenc}    
\usepackage{hyperref}       
\usepackage{url}            
\usepackage{booktabs}       
\usepackage{amsfonts}       
\usepackage{nicefrac}       
\usepackage{microtype}      
\usepackage{graphicx}      
\usepackage{enumitem}      
\usepackage{siunitx}       
\usepackage{etoolbox}
\robustify{\bfseries}
\usepackage{subcaption}

\title{The Eighth Dialog System Technology Challenge}

\author{
  \bf{Seokhwan Kim, Michel Galley, Chulaka Gunasekara, Sungjin Lee,}\\
  \bf{Adam Atkinson, Baolin Peng, Hannes Schulz, Jianfeng Gao, Jinchao Li,}\\
  \bf{ Mahmoud Adada, Minlie Huang, Luis Lastras, Jonathan K. Kummerfeld, }\\
  \bf{Walter S. Lasecki, Chiori Hori, Anoop Cherian, Tim K. Marks,}\\
  \bf{Abhinav Rastogi, Xiaoxue Zang, Srinivas Sunkara, Raghav Gupta}\thanks{Every author has equal contribution. https://sites.google.com/dstc.community/dstc8/}
}

\begin{document}

\maketitle

\begin{abstract}
  This paper introduces the Eighth Dialog System Technology Challenge.
  In line with recent challenges, the eighth edition focuses on applying end-to-end dialog technologies in a pragmatic way for multi-domain task-completion, noetic response selection, audio visual scene-aware dialog, and schema-guided dialog state tracking tasks.
  This paper describes the task definition, provided datasets, and evaluation set-up for each track.
  We also summarize the results of the submitted systems to highlight the overall trends of the state-of-the-art technologies for the tasks.
\end{abstract}

\section{Introduction}
\label{sec:intro}
The Dialog System Technology Challenge (DSTC) is an ongoing series of research competitions for dialog systems.
To accelerate the development of new dialog technologies, the DSTCs have provided common testbeds for various research problems.
The earlier Dialog State Tracking Challenges~\cite{williams2013dialog,henderson2014second,henderson2014third} focused on developing a single component for dialog state tracking on goal-oriented human-machine conversations.
Then, DSTC4~\cite{kim2017fourth} and DSTC5~\cite{kim2016fifth} introduced human-human conversations and started to offer multiple tasks not only for dialog state tracking, but also for other components in dialog systems as the pilot tasks.
From the sixth challenge~\cite{hori2019overview}, the DSTC has rebranded itself as ``Dialog System Technology Challenge'' and organized multiple main tracks in parallel to address a wider variety of dialog related problems.
Most recently, DSTC7~\cite{yoshino2019dialog} focused on developing end-to-end dialog technologies for the following three tracks:
noetic response selection~\cite{dstc7task1,gunasekara2019dstc7}, grounded response generation~\cite{galley2019grounded}, and audio visual scene aware dialog~\cite{alamri2018audio}. 

For the eighth DSTC, we received seven track proposals and went through a formal peer review process focusing on each task's potential for 
(a) broad interest from the research community,
(b) practical impact of the task outcomes, 
and (c) continuity from the previous challenges.
Finally, we ended up with the four main tracks including two newly introduced tasks and two follow-up tasks of DSTC7.
Multi-domain task-completion track (Section~\ref{sec:track1}) addresses the end-to-end response generation problems in multi-domain task completion and cross-domain adaptation scenarios.
NOESIS II (Section~\ref{sec:track2}) explores a response selection task extending the first NOESIS track in DSTC7 and offers two additional subtasks for identifying task success and disentangling conversations.
Audio visual scene-aware dialog track (Section~\ref{sec:track3}) is another follow-up track of DSTC7 which aims to generate dialog responses using multi-modal information given in an input video.
Schema-guided dialog state tracking track (Section~\ref{sec:track4}) revisits dialog state tracking problems in a practical setting associated with a large number of services/APIs required to build virtual assistants in practice.
The remainder of this paper describes the details of each track.

\def\redditNSubreddits{one thousand}
\def\redditNSubscribers{eighty thousand}
\def\redditNTrain{five million}
\def\redditNVal{one million}
\def\redditStart{November 2017}
\def\redditEnd{October 2018}
\def\multiwoz{MultiWOZ}
\newcommand\metalwoz{MetaLWOz}
\section{Multi-Domain Task-Completion Track}
\label{sec:track1}

This track offers two tasks to foster progress in two important aspects of dialog systems: dialog complexity and scaling to new domains. 

\subsection{Task 1: End-to-end multi-domain dialog system}
\label{sec:track1:task1}
Previous work in dialog research communities mainly focuses on individual components in a dialog system and pushes forward the performance of each component. However, the improvement of individual components does not necessarily boost the entire system performance \cite{lee-etal-2019-convlab,INR-074}. The metrics used for an individual component might not be significant for an end-to-end system, and the propagation of error down the pipeline is likely to mitigate the component-wise improvement. With these concerns, recently researchers have taken efforts to create end-to-end approaches \cite{wen2017network,lei2018sequicity}, but it is hard to compare them with conventional methods given the efforts and complexity to combine individual models in conventional approaches. 

To address these concerns, we provide ConvLab (\href{https://github.com/ConvLab/ConvLab}{\texttt{github.com/ConvLab/ConvLab}}) \cite{lee-etal-2019-convlab}, a multi-domain end-to-end dialog system platform covering a range of state-of-the-art models, to reduce the efforts of building and evaluating end-to-end dialog systems. 
Based on ConvLab, participants of the task are to build a dialog system that takes natural language as input, tracks dialog states during the conversation, interacts with a task-specific knowledge base, and generates natural language response as output. There is no restriction on system architectures, and participants are encouraged to explore various approaches ranging from conventional pipeline systems and end-to-end neural approaches.

\subsubsection{Data}
\label{sec:track1:task1:data}

In this task, we consider MultiWOZ  \cite{budzianowski2018multiwoz} dataset, a dialog corpus collected from conversations over multiple domains under the tourist information desk setting. We enhanced the dataset with additional annotation for user dialog acts, which is missing in the original dataset, and included it in ConvLab.

\subsubsection{Evaluation and Results}
\label{sec:track1:task1:evaluation}
\begin{table}[!tb]
    \footnotesize
    \centering
        
        \caption{Task 1 Evaluation Results}
        \setlength{\tabcolsep}{5pt}
        \begin{tabular}{@{}lccccccccccc@{}}
        \toprule
        \multicolumn{1}{c}{} & \multicolumn{4}{c}{Human Evaluation} & \multicolumn{7}{c}{Simulator-based Evaluation} \\
        \cmidrule(r){2-5}\cmidrule(l){6-12}
           Team & Succ. \% & Under. & Appr. & Turns & Succ. \% & Reward & Turns & Prec. & Rec. & F1 & Book \%\\
        \midrule
Best${}^a$ & 68.32 & 4.15 & 4.29 & 19.51 & 88.80 & 61.56 & 7.00 & 0.92 & 0.96 & 0.93 & 93.75  \\
Baseline & 56.45 & 3.10 & 3.56 & 17.54 & 63.40 & 30.41 & 7.67 & 0.72 & 0.83 & 0.75 & 86.37 \\
\bottomrule\\[-2mm]
\multicolumn{12}{@{}l@{}}{\scriptsize ${}^a${The best results for human evaluation and simulator-based evaluation are from  different teams.}}\\[-1mm]
\multicolumn{12}{@{}l@{}}{\scriptsize Metrics: Succ.: success rate, Under.: understanding score, Appr.: appropriateness score, Prec./Rec.: precision/recall of slots prediction.}
        \end{tabular}
        \vspace*{-2mm}
    \label{tab:auto_eval}
\end{table}


Two evaluation metrics are offered in this task:\\[2mm] {\textbf{Simulator-based evaluation}}: The end-to-end user simulator for automatic evaluation is constructed by combining agenda-based user simulator \cite{schatzmann2007agenda}, rule-based NLG and MILU, all of which have been implemented in ConvLab. The evaluation metrics employed include success rate, average reward, and number of turns for each dialog. We also report precision, recall, and F1 score for slot prediction. \\[2mm]
 {\textbf{Crowdworker-based human evaluation}}: With simulator-based automatic evaluation, we filter out low-quality submissions and send the remaining systems to Amazon Mechanic Turk for human evaluation. Crowd-workers communicate with the system via natural language, judge the system and provide ratings based on language understanding correctness, response appropriateness on 5 point scale. Extra metrics including success rate and number of turns are also reported. 

Twelve teams participated in this task. Table~\ref{tab:auto_eval} lists the results for both human evaluation and simulator-based evaluation. A component-wise system with BERT-based NLU model \cite{devlin2019bert}, elaborated rule-based dialog policy and dialog state tracker achieves the best success rate of 88.80\% in simulator-based evaluation.
However, there are discrepancies between human evaluation and simulator-based evaluation. The best system in the human evaluation is based on fine-tuning GPT-2 \cite{radford2019language}. It predicts dialog states, system actions, and responses in an end-to-end fashion, and achieves a success rate of 68.32\%.

\subsection{Task 2: Fast Adaptation Task}
\label{sec:track1:task2}

\vspace*{-2mm}
Neural dialog response generators require very large datasets to learn to output consistent and grammatically correct sentences \cite{vinyals:15,li_deep_2016,wen_latent_2017}. 
This makes it extremely hard to scale out the system to new domains with limited in-domain data, for example, when modeling user responses for a task-oriented chatbot on a narrow domain. 
With this challenge, our goal is to investigate whether sample complexity can decrease with time, \textit{i.e.}, if a dialog system that was trained on a large corpus can learn to converse about a new domain given a much smaller in-domain corpus. 

\vspace*{-1mm}
\subsubsection{Data}
\label{sec:track1:task2:data}
We provide two dialog datasets, where each dialog belongs to exactly one domain.

\textbf{Reddit Dataset} We constructed a corpus of over \redditNTrain{} dialogs from Reddit submissions and comments spanning one year of data. Content is selected from a curated list of \redditNSubreddits{} subreddits using a methodology similar to the DSTC7 sentence generation task \cite{galley2019grounded}. We provide pre-processing code for Reddit data so that all participants work on the same corpus.

\textbf{Goal-Oriented Corpus \metalwoz} We collected 37\,884 goal-oriented dialogs via crowd-sourcing using a \emph{Wizard of Oz} scheme. These dialogs span 47 domains (\textit{e.g.} bus schedule, alarm setting, banking) and are particularly suited for meta-learning dialog models. For each dialog, we paired two crowd-workers, one had the role of being a bot, and the other one was the user. We defined 227 tasks distributed over the domains. 
Note that all entities were invented by the crowd-workers (for instance, the address of a bus stop) and the goal of this challenge is to predict convincing \emph{user} utterances.

\subsubsection{Evaluation and Results}
\label{sec:track1:task2:evaluation}
We evaluate responses by the domain-adapted dialog model using two metrics:\\[2mm]
\textbf{Automatic metrics:} A small set of complete single-domain MultiWOZ \cite{budzianowski2018multiwoz} dialogs is provided to the model, which is then asked to respond to an incomplete dialog. Intents and slot values correctly detected by the baseline NLU (cf. Sec.~\ref{sec:track1:task1}) in the response serve as an indicator that the domain adaptation was successful. We report intent F1 as well as intent+slot F1.

\textbf{Human evaluation:} The model is given a small set of complete dialogs from a held-out \metalwoz{} domain, and is asked to predict a response to an incomplete dialog from the same domain.
Human annotators were asked to judge the appropriateness, informativeness and utility of the responses \cite{galley2019grounded} \emph{given the \metalwoz{} task}, i.e. whether the simulated user tries to complete the task.
Crowd-workers submit pairwise binary preference judgements given dialog context and metric. Pairs are picked using \emph{Multisort} \cite{maystre2017just} and per dialog/metric rankings are aggregated using Copeland's method \cite{copeland51}. We use bootstrapping \cite{hall2009using} over dialog contexts to assess ranking robustness and found it to be stable. Inter-annotator agreement \cite{cohen60-coefficient,callison11findings} is at $\kappa=0.29$. No method outperformed the ground truth.

As a baseline, we provided a retrieval model that relies on FastText \cite{joulin2016fasttext} embeddings of SentencePiece \cite{kudo-richardson-2018-sentencepiece} tokens and only takes into account the given in-domain dialogs. The track received four submissions, all of which surpassed baseline performance on automatic evaluation. As in Task~1 (Sec.~\ref{sec:track1:task1:evaluation}), we find differences in ranking between human and automatic evaluation.

The two best teams use a Transformer \cite{vaswani-transformer} (TeamB) or BiLSTM-based \cite{lstm1997} (TeamA) base model that is fine-tuned on the in-domain dialogs. The BiLSTM-based model is additionally fine-tuned on dynamically sampled Reddit dialogs, while the Transformer model additionally ranks both the observed in-domain dialog responses and the generated response using next sentence classification.

\sisetup{round-mode=places,round-precision=2,per-mode=symbol,detect-all,tight-spacing=true,table-format=1.2,table-column-width=20mm}
\newcolumntype{C}[1]{>{\centering\let\newline\\\arraybackslash\hspace{0pt}}m{#1}}
\begin{table}
    \centering
    \small
    \caption{Fast Adaptation Task Evaluation Results}
    \begin{tabular}{@{}p{20mm}SSSS@{}}
    \toprule
    & \multicolumn2c{Automatic Evaluation} & \multicolumn2c{Human Evaluation} \\\cmidrule(r){2-3}\cmidrule(l){4-5}
    Submission & \multicolumn1c{Intent F1} & \multicolumn1c{Intent \& Slot F1} & \multicolumn1c{Mean Bootstrap Rank} & \multicolumn1c{Final Rank}\\\midrule
    Baseline & 0.515258	& 0.265817 & 3.97 & 4\\
    TeamA &\bfseries 0.786904	&	\bfseries 0.59933 & 3.03 & 3\\
    TeamB &0.644938&	0.483335	& \bfseries 1.01 & \bfseries 1\\
    TeamC &	0.613976&0.418703	& 1.99 & 2\\
    TeamD &	0.549803&0.423359 & 5.00 & 5\\\bottomrule%
    \end{tabular}
    \label{tab:f1}
\end{table}

\pagebreak




\section{NOESIS II: Predicting Responses Track}
\label{sec:track2}

This track is a follow-up to DSTC 7 Track 1, "NOESIS: Noetic End-to-End Response Selection Challenge" \cite{yoshino2019dialog}.
That task considered the next-utterance selection problem in dialogues with two participants and in two domains.
This task extends the challenge in three ways:
(1) conversations with more than two participants;
(2) being able to predict whether a dialogue has solved the problem yet;
(3) handling multiple simultaneous conversations in the same communication channel.
Each of these adds an important aspect of real-world conversations.

\subsection{Task definition}
\label{sec:track2:task}

The primary task is next-utterance selection.
In this problem, each example consists of a partial dialogue and a set of potential messages to come next in the dialogue.
Participants must rank the potential messages plus the possibility that the true next message is not in the set.
We followed the configuration from DSTC 7 track 1, with one hundred options for the next message.
In 20\% of cases the true next message is not in the set.
Participants are also permitted to use certain external knowledge sources in their system.

We also consider three other subtasks that probe specific challenges in dialogue.
Second subtask, a variant of main task in which the conversation context contains a combination of different conversations.
This can occur in settings where a group of people are communicating in the same channel.
To reduce ambiguity about which conversation the next message is part of, we provide the identity of the speaker.
In the third subtask, we consider a task in which the goal is to determine whether the conversation has succeeded in solving the user's problem.
Systems must predict the point in the conversation so far at which success or failure occurred or that no conclusion has been reached yet.
As an optional task, we consider a conversation disentanglement problem, in which data from a channel with multiple conversations must be separated into a set of separate conversations.

\subsection{Data}
\label{sec:track2:data}

As in DSTC 7 track 1, two sources of data were considered.
Both are task oriented, but one is much broader in scope and has more data (Ubuntu) while the other is smaller and more focused (Advising).

\begin{figure}
    \centering
    \small
    \begin{tabular}{lll}
        \toprule
        Time & Speaker & Message \\
        \midrule
        12:30 & s$_{0}$ & how can i boost microphone volume? The volume is toooooo low \\
        12:30 & s$_{1}$ & s$_{0}$ , look for a microphone boost in alsamixer \\
        12:30 & s$_{2}$ & s$_{0}$ : type 'alsamixer' into terminal \\
        12:31 & s$_{0}$ & how the heck do i use alsamixer? :P what is microphone ? \\
        12:32 & s$_{0}$ & how do i choose volume on input s$_{2}$ ? \\
        12:33 & s$_{2}$ & s$_{0}$ : arrow keys up and down \\
        12:33 & s$_{0}$ & s$_{2}$ , yes i understand that. But wich one of those things am i supposed to choose ? \\
        12:33 & s$_{2}$ & s$_{0}$ : you wanted input, right? \\
        12:34 & s$_{0}$ & s$_{2}$ , yes. But i there is no way i can turn that up. :S \\
        12:34 & s$_{2}$ & s$_{0}$ : press tab to go over to capture, then turn it up \\
        12:34 & s$_{0}$ & aha :) thanks \\
        \bottomrule
    \end{tabular}
    
    \vspace{5mm}
    \begin{tabular}{ll}
        \toprule
        Speaker & Message \\
        \midrule
            Student & Hello! \\
            Advisor & Hi! \\
            Student & I am currently trying to figure out what courses to take next semester. \\
            Student & Could you suggest any? \\
            Advisor & Let me see. Give me a minute to go over your transcript \\
            Advisor & Can you tell me what your preferences are? \\
            Student & Of course! I am interested in Computer Science, video game design is something that has always \\
                    & been interesting for me. \\
            Advisor & Eecs 280 I should a prerequisite for most computer science classes, including game design \\
            Student & Okay yeah I will take that course. Do you know of any other prerequisites for game design? \\
            Advisor & Eecs 281 is also necessary, and unfortunately you can't take both 280 and 281 in the same \\
                    & semester. \\
            Advisor & You should take Eecs 203 as that is also a prerequisite for most Eecs classes \\
            Student & Okay thanks for the info! Are both EECS 203 and EECS 280 project based? \\
            Advisor & 280 is all project based and 203 is not, but don't let that fool you. Many students say 203 is \\
                    & harder than 280 \\
            Student & Oh wow okay so do you think that taking them both in the same semester will be manageable? \\
            Advisor & If you have a good grasp of probability and combinations it I should perfectly manageable \\
        \bottomrule
    \end{tabular}
    \caption{\label{fig:track1}
    Examples of data in NOESIS II track: new dialogues from Ubuntu (top) and prior dialogues from Advising (bottom).}
\end{figure}

\paragraph{Ubuntu}
A new set of disentangled Ubuntu IRC dialogs was provided for this challenge based on recent work \cite{acl19disentangle}.
These are derived from the raw Ubuntu logs directly, not from any prior corpus.
The dataset consists of multi-party conversations extracted from the Ubuntu IRC channel.\footnote{
\url{https://irclogs.ubuntu.com/}
}
A typical dialog starts with a question that was asked by one participant, and then other participants respond with either an answer or follow-up questions that then lead to a back-and-forth conversation.
In this challenge, the context of each dialog contains at least three messages between the participants.
The next turn in the conversation is guaranteed to be from one of the participants who has spoken so far.

\paragraph{Advising}
This dataset contains two party dialogues that simulate a discussion between a student and an academic advisor.
The purpose of the dialogues is to guide the student to pick courses that fit not only their curriculum, but also personal preferences about time, difficulty, areas of interest, etc.
The conversations used are the same as those used in DSTC 7 task 1 \cite{yoshino2019dialog}.
They were collected by having students at the University of Michigan act as the two roles using provided personas.
Structured information in the form of a database of course information was provided, as well as the personas (though at test time only information available to the advisor was provided, i.e. not the explicit student preferences).
The data also includes paraphrases of the sentences and of the target responses.

\subsection{Evaluation and Results}
\label{sec:track2:evaluation}

The main task and the second subtask used Recall@k (k=1,10) and mean reciprocal rank (MRR) as the evaluation metrics, following DSTC 7 track 1. The teams were ranked using the mean of recall at 10 and MRR.
The third subtask used accuracy, precision, recall, and f-score which indicates the model's ability to correctly identify whether the dialog task has succeeded or not.

\label{sec:track2:results}
We received 20 submissions from 17 teams. Tables \ref{tab:result_st1_track2} and \ref{tab:result_st2_3_track2} show the performances of the top 3 teams for main task and subtasks respectively. The best performing team (Team 15) of the main task used the BERT \cite{devlin2019bert} and RoBERTa \cite{liu2019roberta} models and fine-tuned the models on the provided in-domain dialogs.

\begin{table}[ht]
\centering 
\caption{\footnotesize{Results of the top 3 performers in Track 2 - main task (subtask 1)}}
\begin{tabular}{|l|ccc||l|ccc|}
\toprule 
    \multicolumn{4}{|c||}{Ubuntu} & \multicolumn{4}{|c|}{Advising}\\\midrule

    Team & Recall@1 & Recall@10 & MRR
    & Team & Recall@1 & Recall@10 & MRR  \\
    \midrule
    15 & 0.761 & 0.979 & 0.848 & 17 & 0.564 & 0.878 & 0.677
    \\
    12 & 0.719 & 0.976 & 0.819 & 15 & 0.306	& 0.762 & 0.455 \\
    5 & 0.663 &	0.974 & 0.786 & 13 & 0.254 & 0.69	& 0.401\\
    \bottomrule
\end{tabular}
\label{tab:result_st1_track2}
\end{table}

\begin{table}[ht]
    \caption{Results of the top 3 performers in Track 2 - Subtask 2 and 3}
    \begin{subtable}{.5\linewidth}
      \centering
        \caption{Subtask 2 - Ubuntu }
        \begin{tabular}{|c|ccc|}
        \toprule
            Team & Recall@1 & Recall@10 & MRR\\\midrule
            15 & 0.706 & 0.957 & 0.799 \\
            13 & 0.596 & 0.904 & 0.707 \\
            3 & 0.505  & 0.834 & 0.621 \\
            \bottomrule
        \end{tabular}
    \end{subtable}%
    \begin{subtable}{.5\linewidth}
      \centering
        \caption{Subtask 3 - Advising }
        \begin{tabular}{|c|cccc|}
        \toprule
            Team & Accuracy & Precision & Recall & F1\\\midrule
            15 & 0.802 & 0.832 & 0.802 & 0.817 \\
            3 & 0.802 & 0.832 & 0.802 & 0.817\\
            13 & 0.662 & 0.707 & 0.687 & 0.697\\
            \bottomrule
        \end{tabular}
    \end{subtable} 
\label{tab:result_st2_3_track2}
\end{table}

\section{Audio Visual Scene-Aware Dialog Track}
\label{sec:track3}
The goal of building an automated system that can converse with humans about video scenes via natural dialogs is a challenging research problem that lies at the intersection of natural language processing, computer vision, and audio processing. As seen at DSTC6 and DSTC7, end-to-end dialog modeling using paired input and output sentences is a way to reduce the cost of data preparation and system development to generate reasonable dialogs in many situations. Such end-to-end approaches have been shown to better handle flexible conversations by enabling model training on large conversational datasets \cite{hori2019overview, yoshino2019dialog}. In the field of computer vision, interaction with humans about visual information has been explored in {\em visual question answering} (VQA) by~\cite{VQA} and {\em visual dialog} (VisDial) by~\cite{visdial_rl}.
The state of the art in video description uses multimodal fusion
to combine different input modalities (feature types), such as spatiotemporal motion features and audio features
\cite{hori2017attention}. 
Since the recent revolution of neural network models allows us to combine different modules into a single end-to-end differentiable network, this framework allows us to build scene aware dialog systems by combining dialog and multimodal video description approaches.
That is, we can simultaneously use video features and user utterances as input to an encoder-decoder-based system whose outputs are natural-language responses. 
To advance research into multimodal reasoning-based dialog generation, we developed the Audio Visual Scene-Aware Dialog (AVSD) dataset and proposed the AVSD challenge in DSTC7. The goal was to design systems to generate responses in a dialog about a video, given the dialog history and audio-visual content of the video. The winning system of the challenge applied hierarchical attention mechanisms to combine text and visual information, yielding a relative improvement of 22\% in the human rating of the output of the winning system vs. that of the baseline system.  
This suggests that there is perhaps significantly more potential in store for advancing this new research area. Toward this end, we propose a second edition of our AVSD challenge in DSTC8.

\subsection{Task definition}
In this track, the system must generate responses to a user input in the context of a given dialog.
 The target of both VQA and VisDial was \emph{sentence selection} based on information retrieval. For real-world applications, however, spoken dialog systems cannot simply select from a small set of pre-determined sentences. Instead, they need to immediately output a response to a user input. For this reason, in this track we focus on \emph{sentence generation} rather than sentence selection. In this track, the system's task is to use a dialog history (the previous rounds of questions and answers in a dialog between user and system) and (optionally) a brief video script, plus (in one version of the task) the visual and audio information from the input video, to answer a next question about the video. The detailed task description is shown at the github page of DSTC8 AVSD\footnote{https://github.com/dialogtekgeek/DSTC8-AVSD}.

\subsection{Data and Baseline System}
\label{avsd-data}
We collected (in~\cite{alamri2018audio}) text-based dialogs about short videos from the Charades dataset\footnote{\url{http://allenai.org/plato/charades/}}~\cite{sigurdsson2016hollywood}, which consists of untrimmed and multi-action videos along with a brief script for each video. 
The data collection paradigm for dialogs was introduced in \cite{Alamri_2019_CVPR}.
In our audio visual scene-aware dialog case, two parties had a discussion about events in a video.
One of the two parties played the role of an answerer who had already watched the video and read the video script.
The answerer answered questions asked by their counterpart, the questioner.
The questioner was not allowed to watch the video but was able to see three frames of the video (the first, middle, and last frames) as static images.
The two parties had 10 rounds of Q and A, in which the questioner asked about the events that happened in the video.
At the end, the questioner summarized the events in the video as a video description. This downstream task incentivized the questioner to collect useful answers for the video description.

The baseline system and an additional submitted system featuring encoder-decoder models using multimodal fusion are described in~\cite{hori_2019_ICASSP}. Detailed results from all models on the DSTC7 challenge, including additional techniques and data set details, were reported in \cite{alamri@DSTC7}.

\subsection{Evaluation}
The automatically generated answers are evaluated by comparing with the 6 ground truth sentences (one original answer and 5 subsequently collected answers).
We used the MS COCO evaluation tool for objective evaluation of system outputs\footnote{https://github.com/tylin/coco-caption}.
The supported metrics include word-overlap-based metrics such as BLEU, METEOR, ROUGE\_L, and CIDEr.
We also collected human ratings of the responses of each system using a 5-point Likert Scale, where humans rated system responses given a dialog context as: 5 (very good), 4 (good), 3 (acceptable), 2 (poor), or 1 (very poor).

\subsection{What We Learned from DSTC7}
AVSD at DSTC7 was the first attempt to combine end-to-end conversation and end-to-end multimodal video description models into a single end-to-end differentiable network to build scene-aware dialog systems.
Most systems employed an LSTM, Bi-LSTM, or GRU encoder/decoder. Some systems used hierarchical and attention frameworks. Furthermore, several additional techniques were introduced to improve the response quality, such as MMI and Episodic Memory Module \cite{alamri@DSTC7}. 
The best system applied hierarchical attention mechanisms to combine text and visual information, yielding an improvement of 22\% in human ratings compared to the baseline system. The language models trained from QA (without video or audio) also performed strongly despite the lack of multimodal information.

After the AVSD challenge at DSTC7, \cite{Alamri_2019_CVPR} reported the performance of sentence selection (as opposed to sentence generation, which was used in this AVSD challenge) using the AVSD dataset. 
In the paper,
\texttt{Question (Q)},
\texttt{V (Video)},
\texttt{Dialog History (DH)}, and
\texttt{Audio (A)} were fused.
The addition of audio features generally improves model performance (\texttt{Q+V} to \texttt{Q+V+A} being the exception). Interestingly, the model performance 
improves even more when combined with dialog history and video features (\texttt{Q+DH+V+A}) for some metrics, indicating that audio signals still provide complementary knowledge to the video signals despite their close relationship.

Further, it is found that the best performance is achieved when including text features extracted from the available summary (video script). Using such manual descriptions improves the performance of all systems. However, such summaries are unavailable in the real world, posing challenges during deployment. Recently, \cite{hori_2019_Interspeech} proposed an approach to transfer the power of a teacher model that was trained using summaries to a student model that does not have access to summaries at test time.

\subsection{DSTC8 Results}
The AVSD Task received 27 system submission from 12 teams.
The best system applied "Fine tuned seq-to-seq model with GPT-2 embedding".
Table \ref{tab:result4avsd} shows 
the evaluation results for the baseline and best systems at DSTC7 and DSTC8 in terms of human rating.
\begin{table}[h]
\centering 
\caption{\footnotesize{Performance comparison between the baseline and the best system.}}
\label{tab:result4avsd}
\begin{tabular}{c|c|c|c|c|c}
\hline
Task & System & BLEU-4 & METEOR & CIDEr & Human rating\\
\hline
\hline
      & Baseline &  0.309 &	0.215 & 0.746 & 2.848\\
\cline{2-6}
DSTC7 & Best & 0.394 & 0.267 & 1.094 & 3.491\\ 
\cline{2-6}
& Human & -  & - &  - & 3.938 \\ 
\hline
       & Baseline &  0.289 &	0.21 & 0.651 & 2.885\\
\cline{2-6}
DSTC8 & Best & 0.442 & 0.287 & 1.231 & 3.934\\ 
\cline{2-6}
      & Human & -  & - &  - & 4.000 \\ 
\hline
\end{tabular}
\end{table}

\subsection{Summary}
We followed up the natural language generation task for Audio Visual Scene-Aware Dialog (AVSD) in DSTC8.
This is the attempt to combine end-to-end conversation and end-to-end multimodal video description models into a single end-to-end differentiable network to build scene-aware dialog systems.
The language models trained from QA and video description are still strong approaches 
but the quality of the results obtained using text only models and multimodal fusion models are almost comparable at this task.
The power to predict the objects and events in the video has been improved and answer the questions more correctly.
Future work includes an exploratory research on reasoning features in response to questions.

\section{Schema-Guided Dialogue State Tracking Track}
\label{sec:track4}
Today's virtual assistants such as the Google Assistant, Alexa, Siri, Cortana, etc. help users accomplish a wide variety of tasks including finding flights, searching for nearby events, surfacing information from the web etc. They provide this functionality by offering a unified natural language interface to a variety of services and APIs from the web.  Building such large scale assistants offers many new challenges such as supporting a large variety of domains, data-efficient handling of APIs with similar functionality and reducing maintenance overhead for integration of new APIs among others. Despite tremendous progress in dialogue research, these critical challenges have not been sufficiently explored, owing to an absence of datasets matching the scale and complexity presented by virtual assistants. To this end, we created the Schema-Guided Dialogue (SGD) dataset, a large-scale corpus of over 18K multi-domain task-oriented conversations spanning 17 domains. This track explores the aforementioned challenges on this dataset, focusing on dialogue state tracking (DST). 

\subsection{Task definition}
\label{sec:track4:task}
The dialogue state is a summary of the entire conversation till the current turn. In a task-oriented system, it is used to invoke APIs with appropriate parameters as specified by the user over the dialogue history. The state is also used by the assistant to generate the next actions to continue the dialogue. DST, therefore, is a core component of virtual assistants. Deep learning-based approaches to DST have recently gained popularity. Some of these approaches estimate the dialogue state as a distribution over all possible slot-values \cite{henderson2014,wen2017network}  or individually score all slot-value combinations \cite{mrkvsic2017neural,zhong-etal-2018-global}. Such approaches are, however, hard to scale to real-world virtual assistants, where the set of possible values for certain slots may be very large (date, time or restaurant name) and even dynamic (movie or event name). Other approaches utilizing a dynamic vocabulary of slot values \cite{rastogi2018multi,goel2019hyst} still do not allow zero-shot generalization to new services and APIs \cite{wu-etal-2019-transferable}, since they use schema elements i.e. intents and slots as class labels.

\begin{figure}
    \centering
    \includegraphics[width=0.99\textwidth]{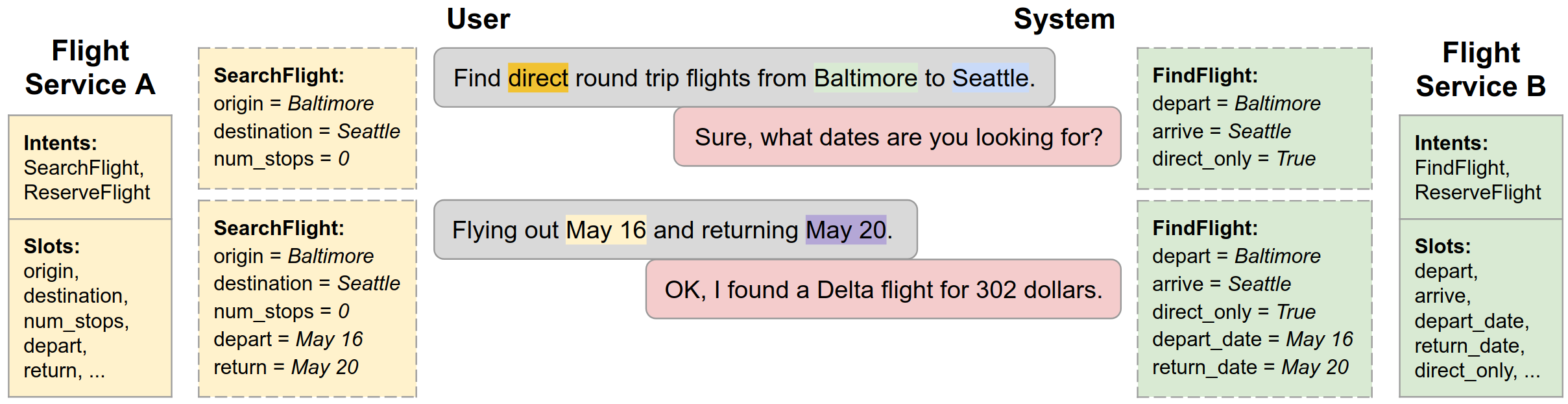}
    \caption{Illustration of Track 4: the dialogue state (dashed edges) for the same dialogue is conditioned on the domain/service schema under consideration (extreme left/right), provided as input.}
    \label{fig:track4-model}
\end{figure}

The primary task of this challenge is to develop multi-domain models for DST with particular emphasis on joint modeling across different services or APIs (for data-efficiency) and zero-shot generalization (for handling new/unseen APIs). This takes the shape of a DST task where the dialogue state annotations are guided by the APIs under consideration. Figure \ref{fig:track4-model} illustrates how the dialogue state representations can be conditioned on the corresponding schema for two different flight services (extreme left and right). In order to generate these schema-guided dialogue state representations, the systems are required to take the relevant schemas as additional inputs. The systems can also utilize the natural language descriptions of slots and intents supported by the APIs to yield distributed semantic representations, which can help in joint modeling of related concepts and generalization to new APIs. In addition, the participants are allowed to use any external datasets or resources to bootstrap their models.

\subsection{Data and Baseline}
\label{sec:track4:data}

\begin{table}[!htb]
    \centering
    \setlength\tabcolsep{4.1pt}
    \def\arraystretch{1.18}
    \resizebox{\textwidth}{!}{
        \begin{tabular}{ l | cc||l | cc ||l | cc } 
        \toprule
            \textbf{Domain} & \textbf{\#Intents}  & \textbf{\#Dialogs} & \textbf{Domain} & \textbf{\#Intents}  & \textbf{\#Dialogs} & \textbf{Domain} & \textbf{\#Intents}  & \textbf{\#Dialogs}\\ \hline
            Alarm & 2 (1) & 37 & Home & 2 (1) & 1027 & Restaurant & 4 (2) & 2755 \\
            Bank & 4 (2) & 1021 & Hotel & 8 (4) & 3930 & RideShare & 2 (2) & 1973\\
            Bus & 4 (2) & 2609 & Media & 4 (2) & 1292 & Service & 8 (4) & 2090\\
            Calendar & 3 (1) & 1602 & Movie & 4 (2) & 1758 & Travel & 1 (1) & 2154\\
            Event & 5 (2) & 3927 & Music & 4 (2) & 1486 & Weather & 1 (1) & 1308\\
            Flight & 8 (3) & 3138 & RentalCar & 4 (2) & 1966\\
            \toprule
          \end{tabular}
 }
    \caption{The number of intents (services in parentheses) and dialogues per domain in the train and dev sets for Track 4. Multi-domain dialogues contribute to counts of each domain.}
    \label{table:track4-domains}
\end{table}

The SGD dataset\footnote{\url{https://github.com/google-research-datasets/dstc8-schema-guided-dialogue}} consists of over 18K annotated multi-domain task-oriented conversations between a human and a virtual assistant. These conversations involve interactions with services/APIs spanning 17 domains (see Table \ref{table:track4-domains}). For most of these domains, SGD contains multiple APIs having overlapping functionalities but different interfaces - common in the real world; it is the first dataset set up this way. The schemas for all services/APIs pertinent to a dialogue, as well as natural language descriptions and other semantic features for a service and its intents and slots, are also included in the dataset. \cite{rastogi2019towards} contains more details about the dataset and the data collection methodology.

With annotations for slot spans, intent, dialogue state and system actions, our dataset is designed to serve as an effective testbed for intent prediction, slot filling, state tracking and language generation, among other tasks in large-scale virtual assistants. Furthermore, the evaluation set is tailored to contain many new services not present in the training set. This helps to quantify the robustness to changes in an API's interface or the addition of new APIs.

We also provide a baseline system \cite{rastogi2019towards}, using user and system utterances and schema element descriptions as inputs to a model based on BERT \cite{devlin2019bert}. The baseline model extends BERT-DST \cite{chao2019bert} by removing all domain-specific parameters, accomplishing zero-shot generalization to new APIs. 

\subsection{Evaluation}
\label{sec:track4:evaluation}
\textbf{Joint goal accuracy}, popular for DST evaluation, is our primary metric for comparison of different approaches, with a modification that uses a fuzzy matching score for non-categorical slots (i.e. slots with large or unbounded sets of possible values) to reward partial matches. For better understanding of the underlying models, we define other auxiliary metrics such as:

\begin{itemize}[leftmargin=*]
    \item \textbf{Active Intent Accuracy:} Fraction of user turns for which the active intent is predicted correctly.
    \item \textbf{Requested Slot F1:} Macro-averaged F1 score for slots requested by the user over all valid turns.
    \item \textbf{Average Goal Accuracy:} Average accuracy of predicting the slot assignments for a turn correctly. Like the joint goal accuracy, this also uses a fuzzy matching score for non-categorical slots.
\end{itemize}

\subsection{Results}

We received submissions from 25 teams. Table \ref{tab:track4-results} lists the results for the top 3 teams (determined by joint goal accuracy) and the baseline system. The evaluation set includes three new domains - ``Messaging", ``Payment" and ``Trains", in addition to having a few unseen APIs for some of the domains present in training and dev sets. We observe that the submitted models are able to generalize well to new APIs and domains. Most of the submitted models make use of pre-trained models like BERT \cite{devlin2019bert}, XLNet \cite{yang2019xlnet} etc. to generalize to unseen domains and APIs.

We also observe a higher joint goal accuracy metric than reported on other public datasets. This is because our dataset excludes the slots for APIs not under consideration in the current turn from the dialogue state for multi-domain dialogues, as opposed to other datasets which include slots for all domains and APIs present over the dialogue history. Thus, in our setup, an incorrect dialogue state prediction for a service only penalizes the joint goal accuracy metric for the turns in which that service is under consideration by the user or the system. Further, our fuzzy matching score rewards partial matches for non-categorical slots, leading to still higher joint and average goal accuracy values.

\begin{table}
    \centering
    \small
    \caption{Evaluation Results for Schema-Guided State Tracking track}
    \begin{tabular}{l|c|c|c|c}
    \toprule
    Team & Joint Goal Accuracy & Avg Goal Accuracy & Active Intent Accuracy & Requested Slots F1 \\
    \midrule
    Baseline & 0.254 & 0.560 & 0.906 & 0.965 \\ 
    Team 9 & 0.865 & 0.971 & 0.948 & 0.985 \\
    Team 14 & 0.773 & 0.922 & 0.969 & 0.995 \\
    Team 12 & 0.738 & 0.920 & 0.926 & 0.995 \\
    \toprule
    \end{tabular}
    \label{tab:track4-results}
\end{table}

\section{Conclusions}
\label{sec:conclusion}
This paper summarizes the tracks of the eighth dialog system technology challenges (DSTC8).
Multi-domain task-completion track offered two sub-tasks: end-to-end multi-domain dialog task and fast adaptation task.
NOESIS II track extended the response selection task of DSTC7 with new datasets with multi-party dialogs and two additional subtasks.
Audio visual scene-aware dialog track explored further improvements from its first edition on DSTC7 with a new test dataset.
Schema-guided dialog state tracking track introduced a new dialog state tracking task from a practical perspective.
All the datasets and resources introduced for every track will still be publicly available after the challenge period to support future dialog system research.

\bibliographystyle{unsrt}
\bibliography{dstc8,refs_AVSD}

\end{document}